\documentclass{article}

\usepackage[utf8]{inputenc} 
\usepackage[T1]{fontenc}    
\usepackage{hyperref}       
\usepackage{url}            
\usepackage{booktabs}       
\usepackage{amsfonts}       
\usepackage{nicefrac}       
\usepackage{microtype}      
\usepackage{amsmath}
\usepackage{graphicx}
\usepackage{algorithm,algorithmic}

\title{Provably robust deep generative models}

\author{
  Filipe Condessa \\
  Bosch Center for Artificial Intelligence\\
  Pittsburgh, PA, USA \\
  \texttt{filipe.condessa@us.bosch.com}\\\\
  J. Zico Kolter \\
  Carnegie Mellon University \& \\Bosch
  Center for Artificial Intelligence\\
  Pittsburgh, USA \\
  \texttt{zico.kolter@us.bosch.com}}

\def \z {\mathbf{z}}
\def \x {\mathbf{x}}
\def \v {\mathbf{v}}
\def \E{\mathbb{E}}
\def \W{\mathbf{W}}

\begin{document}

\maketitle

\begin{abstract}
Recent work in adversarial attacks has developed \emph{provably} robust methods for training deep neural network classifiers.  However, although they are often mentioned in the context of robustness, deep \emph{generative} models themselves have received relatively little attention in terms of formally analyzing their robustness properties.  In this paper, we propose a method for training provably robust generative models, specifically a provably robust version of the variational auto-encoder (VAE).  To do so, we first formally define a (certifiably) robust lower bound on the variational lower bound of the likelihood, and then show how this bound can be optimized during training to produce a robust VAE.  We evaluate the method on simple examples, and show that it is able to produce generative models that are substantially more robust to adversarial attacks (i.e., an adversary trying to perturb inputs so as to drastically lower their likelihood under the model).
\end{abstract}

\section{Introduction}


Adversarial attacks on deep learning system, the observation that deep classifiers are particularly susceptible to small perturbations of inputs which can drastically change class labels, have risen to prominence in recent years \cite{szegedy2013intriguing,goodfellow2014explaining}.  Although several methods have been proposed for defending classifiers again adversarial attacks, many of the more heuristic approaches have proven ineffective \cite{athalye2018obfuscated}.  Indeed, at this point the two primary classes of defenses that have proven effective are adversarial training \cite{goodfellow2014explaining,kurakin2016adversarial,madry2017towards}, which is empirically effective but comes with no guarantees, and provably robust training \cite{wong2018provable,raghunathan2018certified,mirman2018differentiable}, which provides guarantees on the robustness of the resulting classifier but which empirically performs slightly worse.

A common theme among many heuristic defenses is  using generative models to aid in the defense adversarial attacks, i.e., by attempting to ``project'' an input into the latent space of a generative model \cite{ilyas2017robust}, or use the generative model to evaluate likelihood \cite{schott2018towards}.  Such approaching are intuitively appealing: if we have a generative process of our data, then we should be able to account for small perturbations by either flagging them as out of distribution or finding closer examples that do exist within the distribution.  However as mentioned above, despite this potential, such advantage is often not borne out in practice, and generative-model-based approaches do not illustrate a demonstrable advantage over traditional classifiers with adversarial training.  We surmise that this is due at least partially to the fact that deep generative models \emph{themselves} are subject to adversarial attacks.  Indeed, although the majority of work in adversarial attacks has focused on classification systems, recent work has also shown that generative models are equally susceptible to such attacks \cite{athalye2018obfuscated}.  And although some preliminary work exists on using heuristic adversarial training to robustly train generative models \cite{kos2018adversarial}, the \emph{provably robust} approach has no analogue in generative models.

In this paper, we propose the first approach to training \emph{provably robust} deep generative models.  Specifically, we develop an approach to training a provable robust variational auto-encoder (VAE), a method based upon the so-called evidence lower bound (ELBO) on the likelihood.  Our basic approach is to define a robust analogue of unsupervised maximum likelihood estimation, and then use techniques from certifiable robustness to provide a strict lower bound on the ELBO under adversarial perturbations.  Owing to the terms in the ELBO for VAEs, this lower bound is substantially more complex than in the classification setting (where the goal of the adversary is simply to maximize a linear function of the last layer of the network); key to our analysis is the adversarial optimization problem can be framed as a maximizing a set of separable \emph{convex} functions over unit intervals, which has a simple closed form solution despite being a non-convex problem.

To show the effect of provably robust deep models, we compare the performance of a VAE trained in a regular fashion, i.e. with the ELBO as the objective, with the performance of provably robust VAE (proVAE) where the objective is the lower bound of the ELBO for different values of maximum allowed perturbation (considering perturbations within an $\ell_p$ ball of radius $\epsilon_\textrm{train}$).
This performance is measured by evaluating how adversarial attacks affect the log-likelihood / ELBO of the generative model as the attacks become increasingly stronger - projected gradient descent (PGD) attacks with increasing radius $\epsilon_\textrm{attack}$.
The models that are trained in a robust fashion can better defend against the adversarial attacks (higher log-likelihood at higher values of $\epsilon_\textrm{attack}$ as $\epsilon_\textrm{train}$ increases).
Unsurprisingly, increased of robustness comes at the cost of lower log-likelihood when no perturbations exists, analogous to what accuracy decrease trade-off
present in robustly trained classifiers, and we analyze these trade-off in detail, both in terms of the log-likelihood but also in terms of the quality of samples generated from the model.

\section{Background and preliminaries}

\paragraph*{Adversarial deep learning}
Recent years have seen an increased interest in adversarial attacks on machine learning systems \cite{szegedy2013intriguing,goodfellow2014explaining}.  This interest is motivated both by security considerations for deep learning system \cite{eykholt2018robust}, and by the general extent to which adversarial examples cast doubt upon our understanding of what the classifier may be actually picking up on in images \cite{tsipras2018robustness,ilyas2019adversarial}.  In the classification setting, the most common formulation of adversarial robustness replaces the normal expected loss of a classifier
\begin{equation}
    \mathbb{E}_{\x,y \sim \mathcal{D}} \left [ \ell(h_\theta(\x),y) \right]
\end{equation}
(where $\mathcal{D}$ represents the data distribution over inputs $\x$ and outputs $y$, $h_\theta$ is our network, and $\ell$ is a classification loss function), with a term that evaluates the \emph{worst-case} loss over some perturbation applied to the inputs
\begin{equation}
    \mathbb{E}_{\x,y \sim \mathcal{D}} \left [ \max_{\delta \in \Delta(\x)} \ell(h_\theta(\x + \delta),y) \right]
\end{equation}
where $\Delta(\x)$ denotes the allowable set of perturbations. These inner maximization problems, which are non-convex owning to the deep network, (and the resulting training, which minimizes the loss over $\theta$) are typically solved by one of two approaches.  First, in adversarial training, we empirically find a $\delta$ that maximizes the inner optimization, usually via iterative methods such as projected gradient descent (PDG) \cite{kurakin2016adversarial,madry2017towards}; this works well in practice, but doesn't avoid the possibility that some worse attack exists that cannot be found by PGD.  Second, in provably robust methods, we construct a closed-form \emph{upper bound} on the inner maximization problem
\begin{equation}
\max_{\delta \in \Delta(\x)} \ell(h_\theta(\x + \delta),y) \leq \overline{\mathcal{L}(\x, \Delta(\x), y)}
\end{equation}
where $\overline{\mathcal{L}}$ can be constructed in any number of ways, such as linear programming relaxations and their duals \cite{wong2018provable,wong2018scaling,dvijotham2018dual}, semidefinite relaxations \cite{raghunathan2018certified,raghunathan2018semidefinite}, forward bound propagation using zonotopes \cite{mirman2018differentiable,singh2018fast}, or simple box constraints (known as interval bound propagation) \cite{gowal2018effectiveness}.

In this paper, we will adopt this later approach of construction a provable bound on the loss, but do so in the context of unsupervised generative models rather than supervised classification tasks.  To the best of our knowledge, this is the first such application of provably robust training in the context of generative models.

\paragraph*{Robust generative models} Although the vast majority of work in adversarial attacks and defenses has focused on the classification setting (or to a lesser extent, other supervised settings such as object detection \cite{xie2017adversarial,song2018physical}, or image segmentation \cite{fischer2017adversarial,metzen2017universal}), a smaller body of work has indeed looked at the possibility of extending these attacks to generative models \cite{gondim2018adversarial,kos2018adversarial,creswell2017latentpoison}.

Broadly speaking, these works have focused largely on the attack side of the problem, and less on the defense.  In general we will focus in this paper on likelihood based generative models, i.e., models that (usually, approximately) maximize some likelihood function
\begin{equation}
    \mathbb{E}_{\x\sim \mathcal{D}} [\log p(\x)]
\end{equation}
for $\mathcal{D}$ a distribution over samples $\x$ and $\log p_\theta(\x)$ the likelihood of the sample under the model.
In general, there are two possible classes of attacks on likelihood-based generative model, which we refer to out-of-distribution attacks and into-distribution attacks.  In out-of-distribution attacks, the goal is to take some sample \emph{from} the distribution $\x \in \mathcal{D}$ (where we abuse notation slightly to indicate that this sample has high likelihood under the true distribution) and perturb it slightly to lower its likelihood
\begin{equation}
    \min_{\delta \in \Delta(\x)} \log p_\theta(\x + \delta).
\end{equation}
In into-distribution attacks, the goal is to take some sample $\x \not \in \mathcal{D}$ and perturb it so as to \emph{maximize} its likelihood so that it no longer appears to belong to the modeled distribution
\begin{equation}
    \max_{\delta \in \Delta(\x)} \log p_\theta(\x + \delta).
\end{equation}

Although both classes of attacks are relevant, we focus here on out-of-distribution attacks, as it is a much closer analogue of robust adversarial training in the classification setting.  The into-distribution class of attacks is substantially more challenging (it is difficult, for instance, to capture a distribution of all samples \emph{not} in $\mathcal{D}$), which mirrors the case in classification, where even robust models can sometime mistakenly classify images never seen in the input distribution \cite{}; thus, exploring these classes of attacks are left for future work.

\paragraph*{The variational auto-encoder}
In this work, we will consider the variational auto-encoder (VAE) \cite{kingma2013auto} as the generative model of choice.  This is due to the fact that the VAE explicitly is trained based upon a bound on the log-likelihood, which we further bound in the adverarial setting.  A VAE is trained based upon the so-called evidence lower bound (ELBO) $\mathcal{L}(\x)$, which expresses the probability $p(\x)$ in terms of a latent variable $\z \in \mathbb{R}^k$ and then bounds the likelihood as
\begin{equation}
\log p(\x) = \log \int p(\x|\z)p(\z)d\z \geq \mathbb{E}_{\z \sim q(\z|\x)} [ \log p(\x|\z) ] - \mathrm{KL}(q(\z|\x)||p(\z)) \equiv \mathcal{L}(\x)
\end{equation}
where $q(\z|\x)$ is a so-called variational distribution, that attempts to approximate the posterior $p(\z|\x)$ (for which case the bound is tight), but which does so via a more tractable distribution class.  In the VAE setting, we choose 
\begin{equation}
    q(\z|\x) = \mathcal{N}(\z;\mu_\theta(\x), \sigma_\theta^2(\x)\mathbf{I}), \quad
    p(\x|\z) = \mathcal{N}(\x;g_\theta(\z),\sigma_0^2 \mathbf{ I}), \quad
    p(\z) = \mathcal{N}(\z;0,\mathbf{I})
\end{equation}
where $\mu_\theta(\x)$ and $\sigma_\theta^2(\x)$ are the ``encoder'' networks that predict the mean and variance of the Normal distribution $q$ from the input $\x$, and $g_\theta(\z)$ is the ``decoder'' network that generates a sample in input space given a latent vector $\z$.

Under these assumptions, the ELBO has the explicit form
\begin{equation}
    \mathcal{L}(\x,\theta) = \frac{1}{2} \biggl (\E_{\z \sim \mathcal{N}(\mu(\x),\sigma^2(\x) I)}\left[ \sigma_0^2\|\x - g_\theta(\z) \|_2^2 \right] + 1^T(\log \sigma_\theta^2(\x) - \sigma_\theta^2(\x)) - \|\mu_\theta(\x)\|^2 \biggr ) + c
\end{equation}
where $c$ is a constant.  The encoder and decoder networks are jointly trained to maximize this lower bound
\begin{equation}
    \max_{\theta} \mathbb{E}_{\x \sim D} [ \mathcal{L}(\x;\theta)]
\end{equation}
typically using stochastic gradient descent, where we replace the sampling procedure $\z \sim \mathcal{N}(\mu(\x),\sigma^2(\x) \mathbf{I})$ with the equivalent process $\z = \mu(\x) + \sigma(\x) \cdot \epsilon \sim \mathcal{N}(0,\mathbf{I})$ to ensure that the mean and variance terms can be backpropagated through (this is the so-called \emph{reparameterization trick}.

\section{Provably robust deep generative models}
\paragraph*{Overview}
This section addresses the problem of formulating a robust defense against out-of-distribution attacks on deep generative models.
We achieve such defense by obtaining a provable robust deep generative models that lower-bounds the ELBO for any admissible perturbation.
Let  $\Delta_{\epsilon_\textrm{train}}(\x)$ be the set of admissible perturbations, a $\ell_\infty$ ball of radius $\epsilon_\textrm{train}$ centered around $\x$ (for lightness of notation, we will use   $\Delta_{\epsilon_\textrm{train}}(\x)$ and $\Delta(\x)$ interchangeably). 
We can obtain a lower bound for the ELBO for all the possible perturbations $\delta \in \Delta(\x)$ as 
\begin{equation}
\underline{\mathcal{L}(\x)} \leq \mathcal{L}(\x + \delta) \leq \log( p(\x+\delta))
\end{equation}
This lower bound provides a certificate of robustness of ELBO: the effect on ELBO of any possible perturbation in $\Delta(\x)$ will be lower bound by $\underline{\mathcal{L}}$. 
The optimization of the lower bound $\underline{\mathcal{L}}$ effectively trains the network to be robust to the strongest possible out-of-distribution attack within $\Delta(\x)$ ($\ell_\infty$ ball of radius $\epsilon_\textrm{train}$ around $\x$).

\subsection{Interval bound propagation on ELBO}
In order to lower bound the ELBO, we will have to perform interval bound propagation throughout the layers of $\mu_\theta$, $\sigma_\theta$, and $g_\theta$ such that we can obtain bounds for the propagation of the perturbations on the input space in terms of the ELBO.
This means that we must bound both the KL divergence of the perturbed input $\textrm{KL}(q(\z|\x + \delta) || p(\z)) $ and the expected value of the perturbed conditional log-likelihood  $\sigma_{0}^2\|\x - g_\theta(\z)\|_2^2$.
We do so by performing interval bound propagation on the encoder networks $\mu_\theta$ and $\sigma_\theta$, and on the decoder network $g_\theta$.

\subsubsection{Preliminaries}
We start by illustrating how to propagate lower and upper bounds on the building blocks of the encoder and decoder networks: linear and convolution layers, and monotonic element-wise activation functions.
This allows us to sequentially connect the different interval bounds, from input to output of the network.
For lightness of notation, we will denote the upper bound of $u$ as $\overline{u}$ and the lower bound as $\underline{u}$, these should be considered element-wise when addressing multidimensional entities.
\paragraph*{Linear operators}
Let us consider $\W \v$ a linear operator $\W$ applied to $\v$, and ($\overline{\v}$, $\underline{\v}$) the element-wise upper and lower bounds of $\v$.
We can decompose the linear operator $\W$ in positive and negative operators $\W = \W_+ + \W_-$ such that $\W_+ = \max(\W,0)$ and $\W_- = \min(\W,0)$, where $\max$ and $\min$ correspond to element-wise maximum and minimum.
Then the upper and lower bounds of the linear operator applied to $\v$ are
\begin{align}
\label{eq:linear}
\overline{\W \v} = \W_+ \overline{\v} + \W_- \underline{\v}, \nonumber\\
\underline{\W \v} = \W_+ \underline{\v} + \W_- \overline{\v}.
\end{align}
These bounds hold for \emph{\textbf{convolution layers}} and \emph{\textbf{linear layers}}, due to their linear nature.
\paragraph*{Monotonic functions}
Let $\v^t = h(\v^{t-1})$ denote a monotonic (non-decreasing or non-increasing) function applied element-wise on $\v^{t-1}$.
We can  express the upper and lower bounds of $\v^t$ in terms of $h$ and the upper and lower bounds of $\v^{t-1}$ as follows,
\begin{align}
\label{eq:monotonic}
\overline{\v^{t}} = \max\{ h(\overline{\v^{t-1}}), h(\underline{\v^{t-1}}) \}, \nonumber\\
\underline{\v^{t}} = \min\{ h(\overline{\v^{t-1}}), h(\underline{\v^{t-1}}) \}.
\end{align}
These bounds hold for monotonic activation functions, such as \emph{\textbf{ReLU}} and \emph{\textbf{sigmoid}}.

\paragraph*{$\ell_2$ Norm squared}
The lower and upper bounds of the $\ell_2$ norm squared of $\v$ can be easily obtained by noting that there is an element-wise dependency on the lower and upper bounds of $\v$.
As $\| \v\|_2^2 = \sum_{i=1}^n \v_i^2$, where $\v_i$ denotes the $i$th component of $v$, we can obtain the respective upper and lower bounds as a function of $\overline{\v}$ and $\underline{\v}$ as follows,
\begin{align}
\label{eq:l2}
\overline{\| \v\|^2 } = \sum_{i=1}^n \max \{ \overline{ \v_i }^2 , \underline{\v_i}^2  \},\nonumber \\
\underline{\| \v\|^2 } = \sum_{i=1}^n \min \{ \overline{ \v_i }^2 , \underline{\v_i}^2  \}.
\end{align}

\subsubsection{Interval bound propagation on the encoder networks}
Given the encoder networks $\mu_\theta$, and $\sigma_\theta$, we construct the encoder networks to be a succession of convolutional layers with ReLU activations with the last layer being a fully connected linear layer.
Let us consider a perturbation $\x_i + \delta$ where $ \delta \in \Delta_{\epsilon_\textrm{train}}(x_i)$, where we can have lower and upper bound of the perturbation  as 
\begin{equation}
\overline{\x_i} = \x_i + \epsilon_\textrm{train} \mathbf{1}; \quad \underline{\x_i} = \x_i - \epsilon_\textrm{train} \mathbf{1}.
\end{equation}

With the propagation of the interval bounds for linear and convolution layers in \eqref{eq:linear} and for the activation functions in \eqref{eq:monotonic}, we can bound the output of the encoder network based on the interval bound propagation of $\underline{\x_i}$ and $\overline{\x_i}$ throughout the components of the encoder network,
\begin{align}
\underline{\mu_i} = \min \{\underline{\mu_\theta}(\underline{\x_i}),\underline{\mu_\theta}(\overline{\x_i})), \quad
\overline{\mu_i} = \max \{\underline{\mu_\theta}(\underline{\x_i}),\underline{\mu_\theta}(\overline{\x_i}))\\
\underline{\sigma_i} = \min\{\underline{\sigma_\theta}(\underline{\x_i}),\underline{\sigma_\theta}(\overline{\x_i})), \quad
\overline{\sigma_i} = \max \{\underline{\sigma_\theta}(\underline{\x_i}),\underline{\mu_\theta}(\overline{\x_i})),
\end{align}
where $\mu_i = \mu_\theta(\x_i)$ and $\sigma_i = \sigma_\theta(\x_i)$ are the outputs of the encoder networks.
This results in bounds for the output of the encoder network $\mu_i$, for $\sigma_i$, and for $\log \sigma_i$ (as logarithm and exponential are monotonic non-decreasing) as function of the magnitude $\epsilon_\textrm{train}$ of the perturbation $\Delta(\x_i)$.

\paragraph*{KL divergence}
Given the bounds on the output of the encoder network, we can bound the KL divergence between $\mathcal{N}(\mu_i, \sigma_i \mathbf{I})$ and $\mathcal{N}(0, \mathbf{I})$ as follows.
\begin{align}
\label{eq:KL_bound}
\underline{\textrm{KL}} = - \frac{1}{2} \sum_{j=1}^J (1 + \max\{\log(\overline{\sigma_{i}})^2_j - (\overline{\sigma_{i}})^2_j ,\log(\underline{\sigma_{i}})^2_j - (\underline{\sigma_{i}})^2_j \} - (\min\{(\overline{\mu_i})^2_j, (\underline{\mu_i})^2_j\} ), \nonumber \\
\overline{\textrm{KL}} = - \frac{1}{2} \sum_{j=1}^J (1 + \min\{\log(\overline{\sigma_{i}})^2_j - (\overline{\sigma_{i}})^2_j ,\log(\underline{\sigma_{i}})^2_j - (\underline{\sigma_{i}})^2_j \} - (\max\{(\overline{\mu_i})^2_j, (\underline{\mu_i})^2_j\} ),
\end{align}
where $(\mu_i)^2_j$ and $(\sigma_i)^2_j$ denote the $j$th component of the squared mean and covariance of the $i$th sample, as outputed by the encoder networks.
\subsubsection{Interval bound propagation on decoder network}
Now we perform the interval bound propagation through the decoder network, continuing from the bounds on $\mu_i$ and $\sigma_i$ at the end of the encoder networks.
We start by obtaining bounds on the latent variable taking in account the reparameterization trick.

\paragraph*{Reparameterization trick}
The bound on the latent variable using the reparameterization trick follows from the bound for linear operators in \eqref{eq:linear}, as the reparameterization is a linear operator.
Let $\epsilon \sim \mathcal{N}(), \mathbf{I})$, $\epsilon_+ = \max(\epsilon, 0)$ and $\epsilon_- = \min(\epsilon, 0)$ such that $\epsilon = \epsilon_+ + \epsilon_-$.
The reparameterization trick decouples the randomness from the encoder by expressing the latent variable as $\z_i = \mu_i + \sigma_i \epsilon$.
We have then that the latent variable $\z_i$, after reparameterization trick, can be bound as
\begin{equation}
    \underline{\z_i} = \underline{\mu_i} + \overline{\sigma_i} \epsilon_- + \underline{\sigma_i} \epsilon_+, \quad
    \overline{\z_i} = \overline{\mu_i} + \underline{\sigma_i} \epsilon_- + \overline{\sigma_i} \epsilon_+.
\end{equation}

\paragraph*{Decoder network}
The bounds on the latent variable $\z$ are then propagated through the decoder network $g_\theta$, as it is composed of linear and convolutional layers (linear operators where bounds can be propagated with \eqref{eq:linear}) with ReLU and sigmoid activations (monotonic activation functions where bounds can be propagated with \eqref{eq:monotonic}).
This means that we can have bounds on the output of the decoder network as a function of the bounds on the latent vector $\z_i$, as with the encoder networks.

\paragraph*{Conditional log-likelihood}
The remaining element that remains to be bound is the conditional log-likelihood $\log p(\x_i|\z_i)$
We fix the the diagonal covariance $\sigma_0 \mathbf{I}$ in $ p(\x_i|\z_i) =  \mathcal{N}(\x; g_\theta(\z_i), \sigma_0^2 \mathbf{I})$.
This reduces the problem of bounding the conditional log-likelihood to bounding $ \| \x_i -  g_\theta(\z_i)\|^2$.
Following \eqref{eq:l2}, we can bound it by
\begin{align}
 \overline{\| \x_i -  g_\theta(\z_i)\|^2} =   \sum_{j=1}^M \max \{(\x_i + \epsilon_\textrm{train} \mathbf{1}-\underline{g_\theta(\z_i)})^2, (\x_i - \epsilon_\textrm{train} \mathbf{1} - \overline{g_\theta(\z_i)})^2\}_j, \nonumber \\
 \underline{\| \x_i -  g_\theta(\z_i)\|^2} =   \sum_{j=1}^M \min \{(\x_i + \epsilon_\textrm{train} \mathbf{1}-\underline{g_\theta(\z_i)})^2, (\x_i - \epsilon_\textrm{train} \mathbf{1}-\overline{g_\theta(\z_i)})^2\}_j, 
\end{align}
where we are taking the element-wise $\max$ and $\min$ and summing in $j$ across the elements of $\x$.

\subsection{Lower Bound on ELBO}
Finally, we can combine the upper and lower bounds for the encoder network and decoder networks, and associated lower bounds on the conditional log-likelihood and upper bound on the KL divergence, as the ELBO takes in account the negative of the KL divergence, obtaining the following bound
\begin{align}
\label{eq:elbo_lower}
\underline{\mathcal{L}(\x_i,\theta)} =   \frac{1}{2} \sum_{j=1}^J (1 + \min\{\log(\overline{\sigma_{i}})^2_j - (\overline{\sigma_{i}})^2_j ,\log(\underline{\sigma_{i}})^2_j - (\underline{\sigma_{i}})^2_j \} - (\max\{(\overline{\mu_i})^2_j, (\underline{\mu_i})^2_j\} ) \nonumber  + \\
+ \frac{1}{2\sigma_0^2}\E_{\epsilon\sim\mathcal{N}(0, \mathbf{I})}\sum_{j=1}^M \min \biggl\{(\x_i + \epsilon_\textrm{train} \mathbf{1}-\underline{g_\theta(\underline{\z_i})})^2, (\x_i - \epsilon_\textrm{train} \mathbf{1} - \overline{g_\theta(\underline{\z_i})})^2, \nonumber\\
(\x_i + \epsilon_\textrm{train} \mathbf{1} -\underline{g_\theta(\overline{\z_i})})^2, (\x_i -  \epsilon_\textrm{train} \mathbf{1} -\overline{g_\theta(\overline{\z_i})})^2 \biggl\}_j,
\end{align}
where we have the upper and lower bounds for the encoder networks propagated, and the reparameterization trick as 
\begin{align}
\underline{\z_i} = \underline{\mu_i} + \overline{\sigma_i} \epsilon_- + \underline{\sigma_i} \epsilon_+, \quad \overline{\z_i} = \overline{\mu_i} + \overline{\sigma_i} \epsilon_+ + \underline{\sigma_i} \epsilon_- ,\nonumber \\
\underline{\mu_i} = \min\{\underline{\mu_\theta(\x_i + \epsilon_\textrm{train} \mathbf{1})},\underline{\mu_\theta(\x_i - \epsilon_\textrm{train} \mathbf{1})}\},\nonumber \\
\overline{\mu_i} = \max\{\overline{\mu_\theta(\x_i + \epsilon_\textrm{train} \mathbf{1})},\overline{\mu_\theta(\x_i - \epsilon_\textrm{train} \mathbf{1})}\},\nonumber \\
\underline{\sigma_i} = \min\{\underline{\sigma_\theta(\x_i + \epsilon_\textrm{train} \mathbf{1})},\underline{\sigma_\theta(\x_i - \epsilon_\textrm{train} \mathbf{1})}\},\nonumber \\
\overline{\sigma_i} = \max\{\overline{\sigma_\theta(\x_i + \epsilon_\textrm{train} \mathbf{1})},\overline{\sigma_\theta(\x_i - \epsilon_\textrm{train} \mathbf{1})}\}.\nonumber \end{align}

\paragraph*{Provable robustness}
The resulting lower bound on the ELBO lower bounds the log-likelihood of a perturbed sample $\log p(\x_i+\delta)$, working as a robustness certificate for the perturbation.
This means that if $\underline{\mathcal{L}} \geq \alpha$ with input interval bounds fed into the encoder being $\x_i - \epsilon_\textrm{train} \mathbf{1}$ and $\x_i + \epsilon_\textrm{train} \mathbf{1}$ (an $\ell_\infty$ ball centered in $\x_i$ of radius $\epsilon_\textrm{train}$), this guarantees that $\log p(\x+\delta) \geq \alpha$ for all $\delta: \| \delta\|_\infty \leq \epsilon_\textrm{train}$.

\subsection{Training a provably robust variational auto-encoder by optimizing the lower bound}
We can train the provably robust deep generative model by optimizing the lower bound of the ELBO $\underline{\mathcal{L}}$, corresponding to optimizing the robustness certificate, instead of optimizing the ELBO $\mathcal{L}$ directly.

\begin{algorithm}
\caption{Training a provably robust deep generative model - single epoch}
    \label{alg:algorithm-label}
\textbf{input} $\epsilon_\textrm{train}$, $\mathcal{X} = \{\x_1, \hdots, \x_N\}$\\
\algorithmicfor{$\quad \x_i \in \mathcal{X}$}\\
$(\underline{\x_i},\overline{\x_i}) = (\x_i - \epsilon_\textrm{train} \mathbf{1}, \x_i + \epsilon_\textrm{train} \mathbf{1})$  \hfill\algorithmiccomment{\small initial bounds on input}\\
$(\underline{\mu_i}, \overline{\mu_i}) = (\min\{\underline{\mu_\theta(\underline{\x_i})},\underline{\mu_\theta(\overline{\x_i)}}\}, \max\{\overline{\mu_\theta(\underline{\x_i})},\overline{\mu_\theta(\overline{\x_i)}}\})$ \hfill \algorithmiccomment{\small propagate bounds on output of encoder $\mu_\theta$}\\
$(\underline{\sigma_i}, \overline{\sigma_i}) = (\min\{\underline{\sigma_\theta(\underline{\x_i})},\underline{\sigma_\theta(\overline{\x_i)}}\}, \max\{\overline{\sigma_\theta(\underline{\x_i})},\overline{\sigma_\theta(\overline{\x_i)}}\})$ \hfill \algorithmiccomment{\small propagate bounds on output of encoder $\sigma_\theta$}\\
$\epsilon \sim \mathcal{N}(0, \mathbf{I})$ \hfill\algorithmiccomment{\small draw sample for reparameterization trick}\\
$\underline{\z_i} = \underline{\mu_i} + \overline{\sigma_i} \epsilon_- + \underline{\sigma_i} \epsilon_+, \quad \overline{\z_i} = \overline{\mu_i} + \overline{\sigma_i} \epsilon_+ + \underline{\sigma_i} \epsilon_- $  \hfill\algorithmiccomment{\small propagate bounds on latent variable $\z_i$}\\
$(\underline{g_i}, \overline{g_i}) = ( \min \{ \underline{g_\theta(\underline{\z_i})}, \underline{g_\theta(\overline{\z_i})} \},\max \{ \overline{g_\theta(\underline{\z_i})}, \overline{g_\theta(\overline{\z_i})} \})$\hfill\algorithmiccomment{\small propagate bounds on output of encoder $g_\theta$}\\
\\
compute $\underline{\mathcal{L}(\x_i, \theta)}$ with \eqref{eq:elbo_lower}\\
$\theta \leftarrow \theta + \eta \nabla_\theta\underline{\mathcal{L}}(\x_i, \theta) $\hfill\algorithmiccomment{\small update encoder and decoder networks to maximize lower bound of ELBO}\\
\textbf{endfor}\\
\textbf{output} $\mu_\theta, \sigma_\theta, g_\theta$
\end{algorithm}

\section{Experimental results}
\paragraph*{Dataset and Architectures} 
We illustrate the robustness of the proposed provably robust training approach for deep generative models on two datasets: MNIST image classification~\cite{lecun1998gradient} and CIFAR10~\cite{krizhevsky2009learning}.
For the encoder network architectures we test a network composed of two convolutional layers of 32 and 64 filters and a fully connected layer of 3136 units for MNIST and 4096 units for CIFAR10, with ReLU activations.
The final layer has 50 units and no activation, corresponding to a 50 dimensional lantent space ($\z \in \mathbb{R}^{50}$), and the encoder networks $\mu_\theta$, $\sigma_\theta$ share all the weights except for the final layer of 50 units (in practice $\sigma_\theta$ will output $\log \sigma\theta$).
For the decoder network with test a network composed of a fully connected layer of 3136 units for MNIST and 4096 units for CIFAR10, and two convolutional layers of 64 and 32 filters, with ReLu activations with exception of the last convolutional layer that has a sigmoid activation.
\begin{table}[ht]
    \centering
\caption{ Train and test values for ELBO $\mathcal{L}$ and lower bound for ELBO $\underline{\mathcal{L}}$ in \emph{\textbf{ unperturbed}} data (MNIST and CIFAR) for proVAE models trained robustly with different values of radius $\epsilon_\textrm{train}$ of the $\ell_\infty$ of perturbations. We note that when $\epsilon_\textrm{train} = 0$ training the lower bound of the ELBO in proVAE defaults into training a VAE by optimizing the ELBO.}
    \label{tab:train_test}
    \begin{tabular}{ccc}
    \begin{tabular}{|c|c|c|c|}
    \multicolumn{4}{c}{MNIST}\\
    \hline
        $\epsilon_\textrm{train}$ & $\underline{\mathcal{L}}$ train & $\underline{\mathcal{L}}$ test & $\mathcal{L}$ test \\
        \hline 
         $0.00$ & $-28.54$ & $-28.36 $& $-28.36$ \\
         $0.01$ & $-36.44$ & $-36.06$ & $-32.35$ \\
         $0.02$ & $-39.69$ & $-39.21$ & $-33.97$ \\
         $0.05$ & $-44.79$ & $-44.39$ & $-36.78$ \\
         $0.1$ & $-54.49$ & $-54.49$ & $-43.41$ \\
         \hline
    \end{tabular}
        &&  \begin{tabular}{|c|c|c|c|}
        \multicolumn{4}{c}{CIFAR}\\
        \hline
        $\epsilon_\textrm{train}$ & $\underline{\mathcal{L}}$ train & $\underline{\mathcal{L}}$ test & $\mathcal{L}$ test \\
        \hline 
         $0.00$ & $-76.08$ & $-76.13 $& $-76.13$ \\
         $0.01$ & $-111.26$ & $-111.55$ & $-92.74$ \\
         $0.02$ & $-132.12$ & $-132.14$ & $-97.39$ \\
         $0.05$ & $-193.93$ & $-193.91$ & $-105.17$ \\
         $0.1$ & $-296.52$ & $-296.52$ & $-138.55$ \\
         \hline
    \end{tabular} \\
    \end{tabular}
\end{table}

\begin{figure}
     \begin{tabular}{cc}
     \includegraphics[width=.45\linewidth]{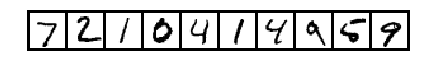} & \includegraphics[width=.45\linewidth]{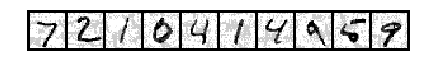}\\
     a) Unperturbed test samples& b) PGD attack on VAE \\
      \includegraphics[width=.45\linewidth]{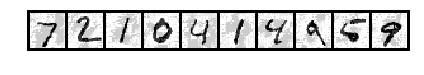}&\includegraphics[width=.45\linewidth]{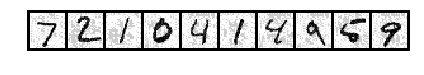} \\
      c) PGD attack on  proVAE $\epsilon_\textrm{train} = 0.01$ &d) PGD attack on proVAE $\epsilon_\textrm{train} = 0.1$ \\
     \end{tabular}
  \caption{Example of adversarial examples on MNIST using the PGD attack with $\epsilon_\textrm{attack} = 0.1$ on the initial unperturbed test samples shown in a).
  }
  \label{fig:mnist_attack}
\end{figure}

\begin{figure}
     \begin{tabular}{cc}
     \includegraphics[width=.45\linewidth]{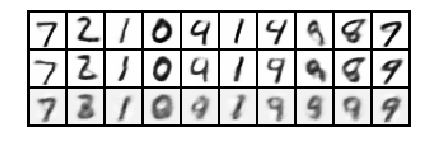} & \includegraphics[width=.45\linewidth]{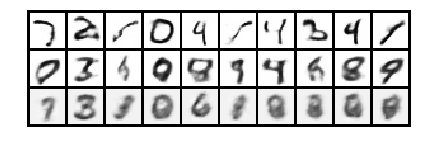}\\
     \end{tabular}
  \caption{Test sample reconstruction of unperturbed data (left) and attacked data (right).
  Top rows corresponds to reconstruction using VAE ($\mathcal{L}=-28.28$ for unperturbed and ${L}=-113.97$ for attacked), middle rows to proVAE $\epsilon_\textrm{train} = 0.01$ ($\mathcal{L}=-31.10$ for unperturbed and $\mathcal{L}=-59.08$ for attacked), and bottom rows to proVAE $\epsilon_\textrm{train} = 0.1$ ($\mathcal{L}=-41.31$ for unperturbed and $\mathcal{L}=-50.06$ for attacked).
  Test samples for unperturbed data come from Figure 1 (a), and for the attacked data corresponds to the PGD attack with $\epsilon_\textrm{attack} = 0.1$ in Figure 1 (b,c,d) for each respective model.}
  \label{fig:mnist_reconstruction}
\end{figure}

\begin{figure}[ht]
\centering
\begin{tabular}{cc}
     \includegraphics[width=.4\linewidth]{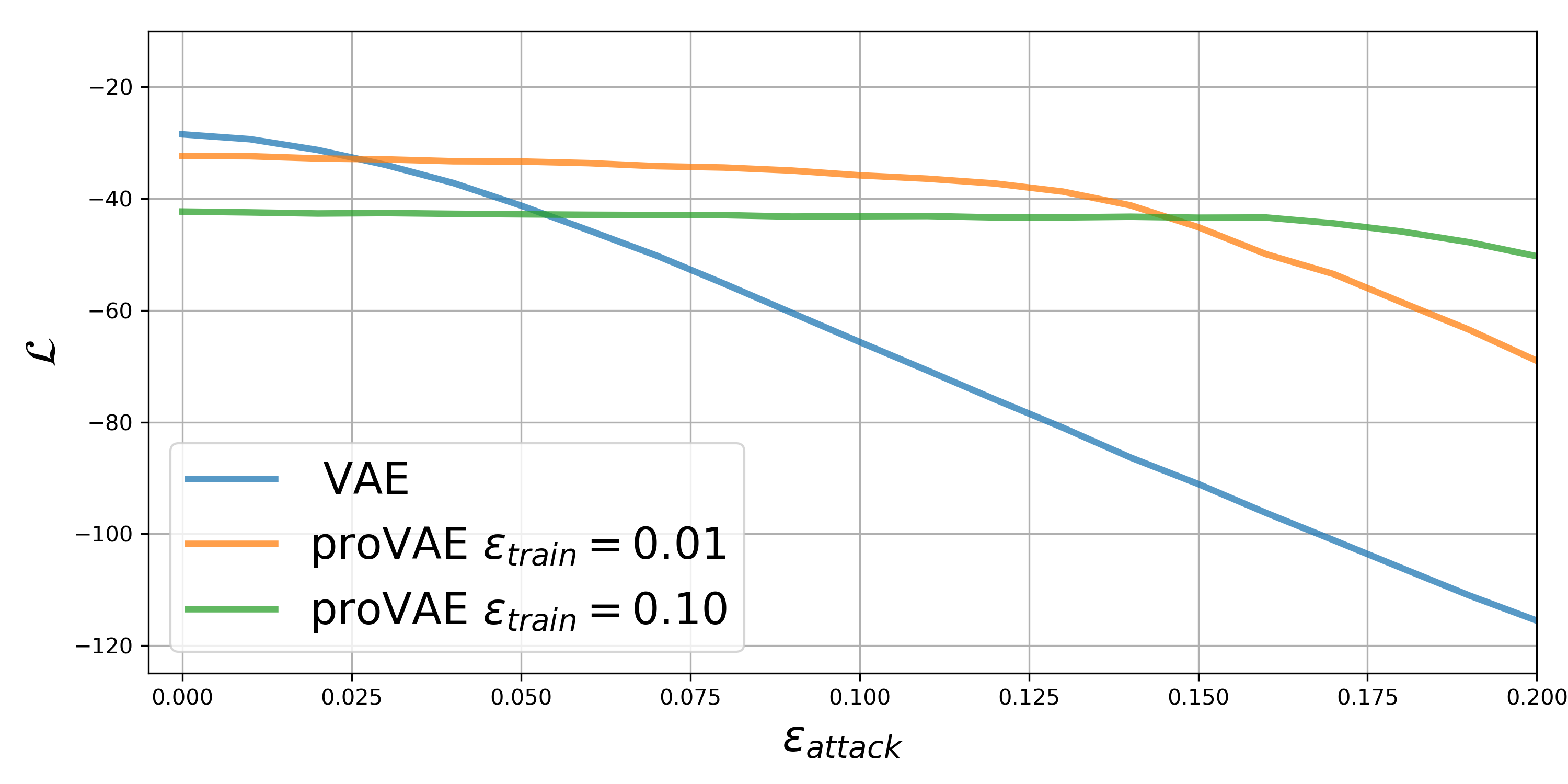}&\includegraphics[width=.4\linewidth]{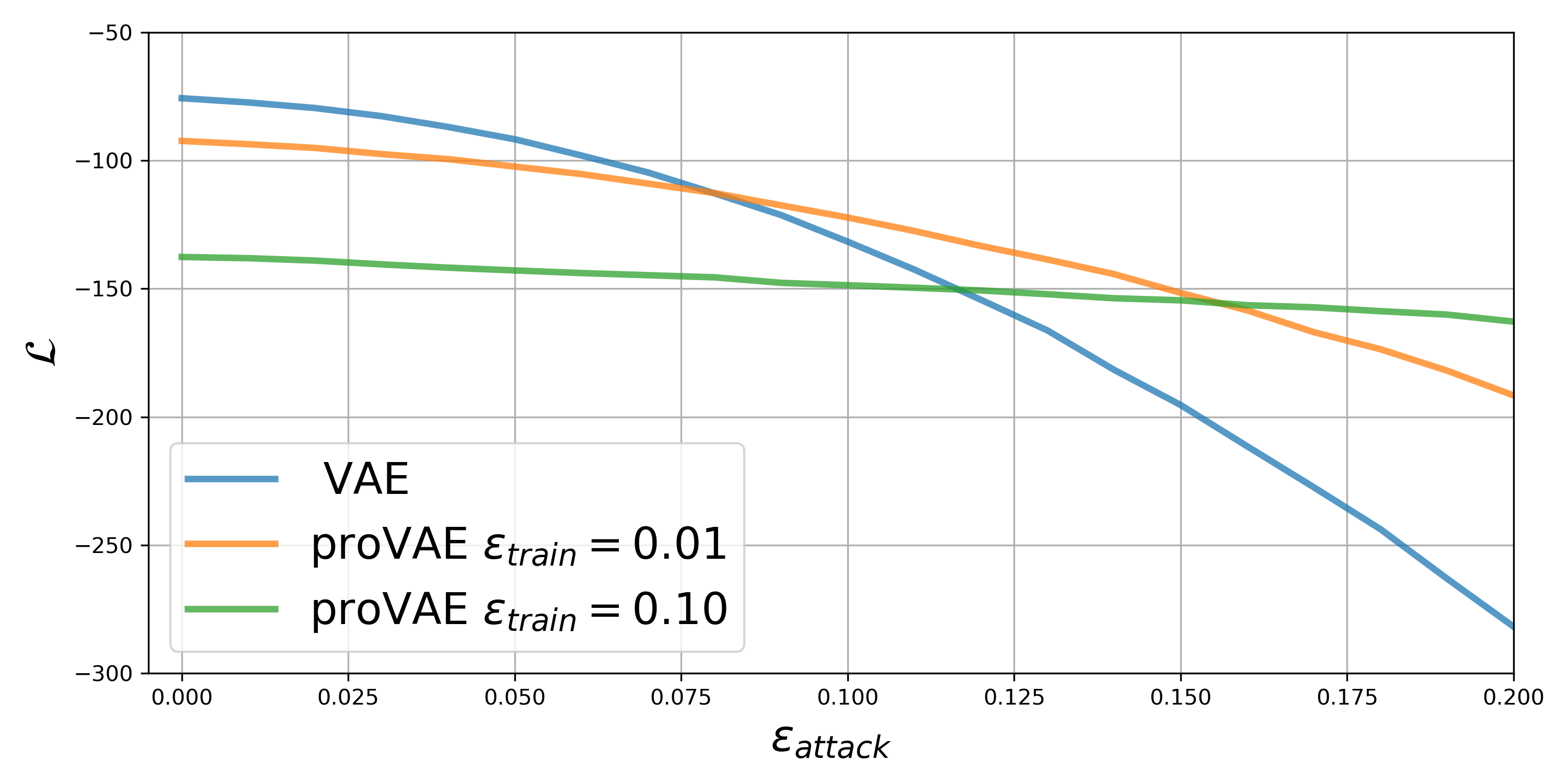}  \\
     MNIST &CIFAR10 
\end{tabular}
  \caption{Effect of adversarial attack (PGD in an $\ell_\infty$ ball of radius $\epsilon_\textrm{attack}$ for $0\leq\epsilon_\textrm{attack} \leq 0.2$) on the ELBO $\mathcal{L}$ for classifiers trained with different values of $\epsilon_\textrm{train}$ (the VAE in bold blue corresponds to $\epsilon_\textrm{train}=0$) on MNIST and CIFAR10. This clearly illustrates the fast decay of $\mathcal{L}$ for the non-robust classifier and how, and how by training a provably robust classifier we can mitigate this decay at the expense of a lower value of $\mathcal{L}$ on unperturbed ($\epsilon_\textrm{attack}$=0) data.}
  \label{fig:elbo_attack}
\end{figure}

\begin{figure}[ht]
\centering
     \begin{tabular}{ccc}
     \includegraphics[width=.25\linewidth]{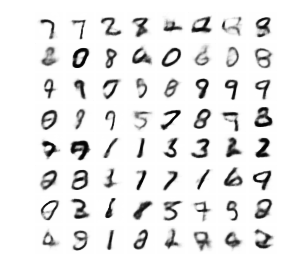}&
     \includegraphics[width=.25\linewidth]{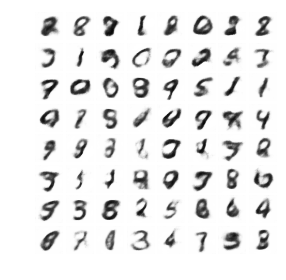}&
     \includegraphics[width=.25\linewidth]{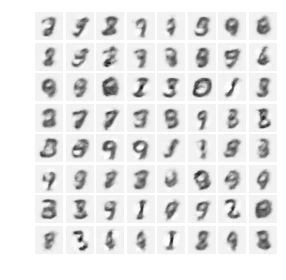}\\
     \vspace{-0.05in}\includegraphics[width=.25\linewidth]{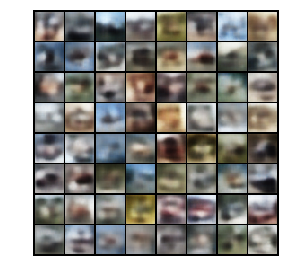}&
    \vspace{-0.05in} \includegraphics[width=.25\linewidth]{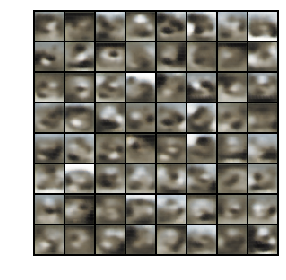}&
     \vspace{-0.05in}\includegraphics[width=.25\linewidth]{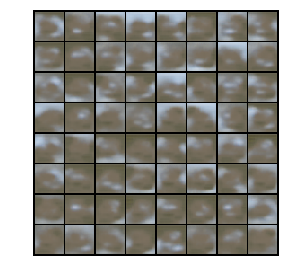}\\
     VAE &  proVAE $\epsilon_\textrm{train} = 0.01$ & proVAE $\epsilon_\textrm{train} = 0.10$ 
\end{tabular}

  \caption{MNIST and CIFAR10 samples drawn from $g_\theta(\epsilon), \epsilon\sim \mathcal{N}(\mathbf{0},\mathbf{I})$.
 The cost of robustness against attacks is present as the samples drawn from more robust models have lower quality, both on MNIST (samples have less sharpness), and on CIFAR (samples degenerate into blurs).}
  \label{fig:rand_samples}
\end{figure}

\paragraph*{Summary of results}
The final training and test lower bounds of ELBO and test ELBO for different data sets and different proVAE models trained with different $\epsilon_\textrm{train}$ are given in Table~\ref{tab:train_test}.
These lower bounds for the ELBO for different $\epsilon_\textrm{train}$ are consistent with the ELBO of those models when subject to a PGD attack of magnitude $\epsilon_\textrm{attack}$ that are given in Figure~\ref{fig:elbo_attack}.
At lower values of $\epsilon_\textrm{attack}$ the models trained at lower $\epsilon_\textrm{train}$ have better performance (higher ELBO as seen in Figure~\ref{fig:elbo_attack}, and generating samples of better quality in Figure~\ref{fig:rand_samples}) than the models trained at higher $\epsilon_\textrm{train}$.
However, as $\epsilon_\textrm{attack}$ increases, the models trained at higher $\epsilon_\textrm{train}$ outperform the models trained at lower $\epsilon_\textrm{train}$.
We point to Figure~\ref{fig:mnist_attack} for examples of PGD attacks ($\epsilon_\textrm{attack} = 0.1$) on VAE and proVAE ($\epsilon_\textrm{train}=0.01$ and $\epsilon_\textrm{train}=0.1$) on MNIST, and to Figure~\ref{fig:mnist_reconstruction} for examples of the reconstructions for the different models of the unperturbed data and from the data perturbed by the PGD attacks.
In Figure~\ref{fig:rand_samples} we find example of samples randomly drawn from those models on MNIST and CIFAR10.
These trade-offs between performance of the nonrobust and robust models as we evaluate them on adversarial and nonadversarial conditions are a clear indication that model robustness comes at a cost, in parallel to the accuracy trade-offs of robustly trained classifiers.
However, training a model to be robust to \emph{small} perturbations ($\epsilon_\textrm{train} \approx 0$) can easily provide more robustness than a traditionally trained mode, at a small cost to the ELBO on nonadversarial conditions.

\section{Conclusion}
In this paper, we presented a method for training provably robust generative models (VAE)  based on defining a robust lower bound to the variational lower bound of the likelihood and optimizing this bound to train a robust generative model (proVAE).
This methodology introduces provable defenses against adversarial attacks into the domain of generative models, namely out-of-distribution attacks where we perturb a sample to lower its likelihood.
The experimental results corroborate the effectiveness of this provable defense, and introduce a new flavor of trade-offs associated with model robustness. 
Whereas this is a significant step towards improving robustness of generative models there are clear directions for improvement.
The most compelling direction for improvement is towards creating defenses against into-distribution attacks, where adversarial attacks can perturb a sample out-of-distribution ($\x \not \in \mathcal{D}$) to make it appear as if it comes from distribution $\x \in \mathcal{D}$.

\end{document}